\documentclass[12pt]{article}

\usepackage{sbc-template}
\usepackage{graphicx,url}
\usepackage[utf8, latin1]{inputenc}
\usepackage[brazil, english]{babel}
\usepackage{wrapfig}

\title{Surveillance Capitalism Revealed: \\Tracing The Hidden World Of Web Data Collection}

\author{Antony Seabra de Medeiros, Luiz Afonso Glatzl Junior, S\'ergio Lifschitz}

\address{Departamento de Inform\'atica \\
        Pontif{\'i}cia Universidade Cat\'olica do Rio de Janeiro (PUC-Rio), Brazil
  \email{\{amedeiros, ljunior, sergio\}@inf.puc-rio.br}
}

\begin{document} 

\maketitle

\begin{abstract}
This study investigates the mechanisms of Surveillance Capitalism, focusing on personal data transfer during web navigation and searching. Analyzing network traffic reveals how various entities track and harvest digital footprints. The research reveals specific data types exchanged between users and web services, emphasizing the sophisticated algorithms involved in these processes. We present concrete evidence of data harvesting practices and propose strategies for enhancing data protection and transparency. Our findings highlight the need for robust data protection frameworks and ethical data usage to address privacy concerns in the digital age.
\end{abstract}

\section{Introduction}
Surveillance capitalism has emerged as a dominant model wherein personal data becomes a pivotal economic commodity in the digital era. Personal data has become a cornerstone of many business models in the digital economy, driving innovation and revenue generation across various industries. For instance, companies like Google and Facebook have built empires primarily based on their ability to collect, analyze, and utilize personal data for targeted advertising. As reported in \cite{zuboff2023age}, 89 percent of the revenues of Alphabet were derived from Google's targeted advertising programs by 2016. The scale of raw-material flows is reflected in Google's domination of the internet, processing over 40,000 search queries every second on average: more than 3.5 billion searches per day and 1.2 trillion searches per year worldwide in 2017.

Similarly, e-commerce platforms like Amazon use personal data to personalize shopping experiences, offering recommendations based on previous purchases, browsing history, and search queries. This not only enhances customer engagement but also significantly boosts sales. In the fitness and health sector, companies like {\it Fitbit} and {\it Strava} collect data on user activities and health metrics, which can be utilized for personalized health and fitness advice and potentially shared with insurance companies or used in healthcare research. These examples underscore the pervasive role of personal data in modern business models, where the value proposition is often based on the depth and breadth of data collected from users.

The primary goal of this study is to gather empirical evidence regarding sharing personal data over the Internet. Specifically, the work aims to investigate data transmission from individual devices to internet-hosted services and discuss potential user concerns regarding privacy and security, contributing to sustainable and ethical data practices in the digital environment. All personal data captured during our case studies are available at \url{https://github.com/antonyseabramedeiros/}.

\section{Surveillance Capitalism and Related Work}
Surveillance capitalism is a term coined by Harvard Business School Professor Shoshana Zuboff in her seminal work \textit{The Age of Surveillance Capitalism} \cite{zuboff2019age}. It describes a new form of capitalism that leverages the analysis of personal data obtained through internet-based platforms and services to create personalized profiles that encapsulate users' behavior and preferences. This data would then be shared across companies and used within a business model to generate predictions about individual users. 

The early internet was a relatively decentralized space, primarily driven by the ideals of free information exchange and user anonymity. In the late 1990s and early 2000s, we saw the rise of tech giants like Google and Facebook (nowadays Meta), which quickly understood that the data generated by users - from search queries to social interactions - were the input for a new business model. Over the years, these companies refined their data capture techniques. Cookies, pixel tracking, and similar technologies were developed to track user behavior online. 

The sophistication of these methods grew alongside advancements in data storage and processing technologies, enabling the accumulation of vast amounts of personal data. This data did not just fuel advertising; it enabled the creation of detailed user profiles, making it possible to predict and influence user behavior. The model evolved beyond advertising as companies began monetizing these insights by offering predictive products to third-party businesses. While enhancing user experience in some aspects, this personalizing capability raised significant privacy concerns. Beyond this, according to \cite{zuboff2022surveillance}, surveillance capitalism is inherently antidemocratic and creates a situation where its expansion leads to democratic instability and the breakdown of institutions, emphasizing the necessity for new public institutions, rights charters, and legal frameworks tailored for a democratic digital era to protect citizens from being exploited through their data.

The implications of the General Data Protection Regulation (GDPR) in the context of increasing surveillance practices driven by Big Data technologies were examined in \cite{andrew2021general}. Its main objective is to analyze the tensions within the GDPR as it balances individual privacy rights with data collection and surveillance realities. The authors argue that while the GDPR has significantly protected personal privacy, it falls short of addressing the broader surveillance risks associated with collecting and trading behavioral data. They highlight that the current legal framework allows for a behavioral futures market with inadequate protections for individuals, suggesting that regulators need to reconsider property rights related to behavioral data to better safeguard citizens against the implications of data commodification and surveillance practices.

On the other hand, Brazil's General Data Protection Law (LGPD) seeks to address the growing concerns over privacy and data protection in a landscape dominated by surveillance practices. Like the GDPR, the LGPD emphasizes individual privacy rights, transparency, and accountability in processing personal data. LGPD legislation was formulated based on the GDPR, according to \cite{martins2020lgpd}, and establishes rules and limits for companies regarding personal data collection, storage, processing, and sharing, especially in digital media, to protect the fundamental rights of freedom, privacy, and the free formation of the personality of each individual. Both regulations impose strict requirements on organizations regarding data collection, consent, and security measures. However, significant differences exist between the two, particularly in their enforcement mechanisms and scope of applicability. While the GDPR grants broader enforcement powers to independent supervisory authorities, the LGPD's regulatory body, the ANPD (National Data Protection Authority), is still developing its capacity and authority. Additionally, the LGPD offers a slightly more flexible approach to legitimate interest processing, potentially creating more leeway for data-driven businesses. Despite these differences, both laws face similar challenges in curbing tech giants' expansive data surveillance practices, suggesting a need for continued evolution of legal frameworks to address the complexities of data commodification and protect individual rights in an era of pervasive digital surveillance.

The case studies presented in \cite{stahl2022surveillance} highlight significant data misuse and privacy violations within the framework of Surveillance Capitalism. The first case focuses on Clearview AI, a company specializing in facial recognition software that has collected billions of images without consent, triggering legal challenges across several European countries. In 2021, France's data protection authority, CNIL, ordered Clearview AI to stop the unlawful processing of biometric data and comply with individuals' rights to access and delete their data. This case exemplifies the practice of data appropriation, where personal data is collected without proper consent or compensation, resulting in significant privacy violations. Clearview AI's actions breached the General Data Protection Regulation (GDPR) by unlawfully processing data and failing to respect individual rights. This highlights the ongoing challenges in regulating companies that profit from unauthorized data collection. These cases collectively underscore the pervasive nature of surveillance capitalism, where companies exploit personal data for commercial gain, often at the expense of individual privacy rights.

The second case involves a 2021 data leak at a New York-based health tracking service provider that exposed the personal information of 61 million users worldwide, including sensitive health data like weight, height, and location. This breach underscores the risks associated with companies using tracking devices' vast collection and storage of health data. It also raises concerns about data privacy and the potential for unauthorized access, especially in light of acquisitions like Google's purchase of Fitbit, which experts argue could lead to monopolistic control and consumer exploitation by combining health data with existing data sets. Another case involving Facebook demonstrates how companies often deceive users about the nature of their services. In 2021, Italy's authorities fined Facebook for misleading users into believing that the service was free without adequately disclosing that their data was being collected and used for commercial purposes. This case illustrates the broader issue of companies not being transparent about how user data is monetized and used for targeted advertising, effectively turning users into products. 

\cite{wu2023slow} investigates the pervasive and often subtle harms caused by online behavioral advertising (OBA) on individuals' lives, framing these harms within slow violence. The authors surveyed 420 participants to identify four primary harms: psychological distress, loss of autonomy, constriction of user behavior, and algorithmic marginalization and trauma. The study highlights how these harms are not just isolated incidents but contribute to broader societal issues of inequality and exploitation. By emphasizing the need for legal recognition of privacy harms and advocating for more comprehensive measures to document and address these issues, the paper calls for a shift in how researchers and policymakers understand and respond to the impacts of OBA on users.

While there are many studies regarding data privacy concerns and regulation proposals, as previously described in \cite{andrew2021general} and \cite{wu2023slow}, we have identified a notable gap concerning the research dedicated to capturing concrete, empirical evidence that elucidates its mechanisms and impacts. To further substantiate our research, we conducted an extensive review of existing literature and studies related to "surveillance capitalism evidence," "empirical evidence," and "data captures" within this context. Despite the abundance of theoretical discussions on surveillance capitalism, our search revealed a significant gap in empirical research that provides concrete evidence of data capture practices by web services. Specifically, we didn't find any studies that methodically document or analyze real-world instances of data interceptions, such as those facilitated by tracking technologies or data aggregation practices. This lack of empirical evidence underscores the necessity of our research, which aims to fill this gap by capturing and analyzing fundamental data interactions to provide tangible insights into the mechanisms of surveillance capitalism. 

\section{Methodology}
Web services on the internet function as intermediaries, enabling devices to communicate and exchange data. A series of data exchanges occurs when a user interacts with a web service through a browser, app, or internet-connected device. These interactions typically involve sending requests from the user's device to the server hosting the web service and receiving responses in return. Our methodology aims to capture these requests and responses to provide a detailed analysis of the interactions.

Data transmission over the Internet is not a direct, point-to-point process. It often traverses through various intermediaries, such as internet service providers (ISPs), routers, and other network nodes. Each point represents a potential place where data can be intercepted or logged. The data transmitted can include the content of a user's interaction with a website and metadata like IP addresses, device identifiers, location data, and browsing history. Web services commonly employ cookies and other tracking technologies to enhance user experience and provide personalized content. However, these technologies also track user behavior across different sites and services. Third-party cookies, in particular, can relay information about a user's browsing habits to external entities, often without explicit consent or awareness of the user.

The potential exposure of personal data extends beyond the primary web service a user interacts with. Data aggregators, which collect information from various sources, including web services, compile extensive profiles of individuals. These profiles can be used for targeted advertising and accessed by other entities, including marketers, insurers, and malicious actors.

The infrastructure required to capture data transmitted from personal devices, such as mobile phones or tablets, involves deploying a proxy server and certificates integrated into the device using a Man-in-the-Middle (MITM) approach. In this configuration, all communication between the device and the internet is intercepted and routed through the proxy server. Message packets can be captured and extracted in plain text format by embedding the server's certificate in the device.

\begin{figure}[ht]
\centering
\includegraphics[width=1\textwidth]{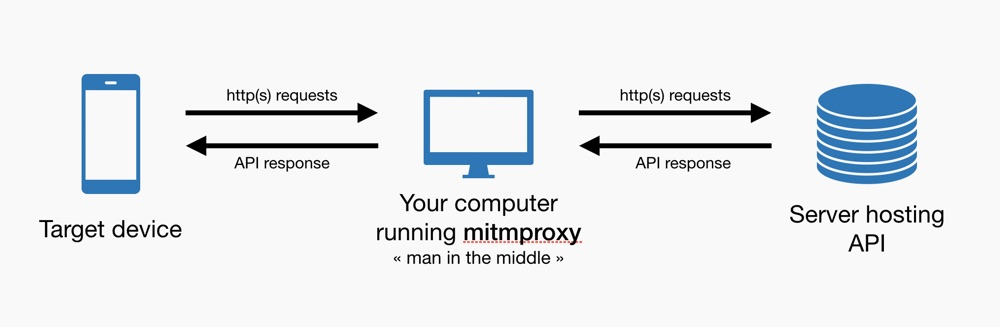}
\caption{Man In The Middle Proxy Interception \cite{mitmserver}}
\end{figure}
%

Setting up the proxy server and configuring the device to utilize it is available at \url{https://github.com/antonyseabramedeiros/}. For this research, we employed the use of the \textit{Mitmproxy} software \cite{mitmproxy}. 
After configuring both the server and the user device, messages transmitted via the HTTP and HTTPS protocols can be intercepted by \textit{Mitmproxy}. The \textit{Mitmweb} utility provides a browser interface for visualizing the intercepted content and building filters to refine the captured data.  This interception reveals all the services receiving messages immediately following a specific action on the test device. 

After connecting a phone to the proxy, users can visit a website or perform a Google search. Mitmproxy captures all outgoing requests and incoming responses during these interactions, displaying them in real time. This allows researchers to observe the direct communication with the intended web service and subsequent requests sent to third-party services, trackers, and ad networks triggered by the initial user action. The captured data highlights the extent of information being shared with multiple entities, often without the user's explicit awareness, thus providing empirical evidence of the data flows underpinning surveillance capitalism.

\section{Case Studies}

This case study examines the extent and mechanisms of data collection during typical web browsing sessions. By navigating through various websites, the study uncovers digital tracking mechanisms. 

\subsection*{Web Navigation}

We start with direct navigation to the website \textit{samsung.com}. This approach is chosen to monitor and analyze the digital footprint left by such navigation, especially the subsequent communication with external services. Upon completing interactions within \textit{samsung.com}, our analysis focuses on the external services contacted due to this initial visit. Specifically, we want to identify and list the entities outside of \textit{samsung.com} that were accessed, as indicated by the inclusion of \textit{samsung.com} in the payload of the network packets. 

\begin{figure}[ht]
\centering
\includegraphics[width=1\textwidth]{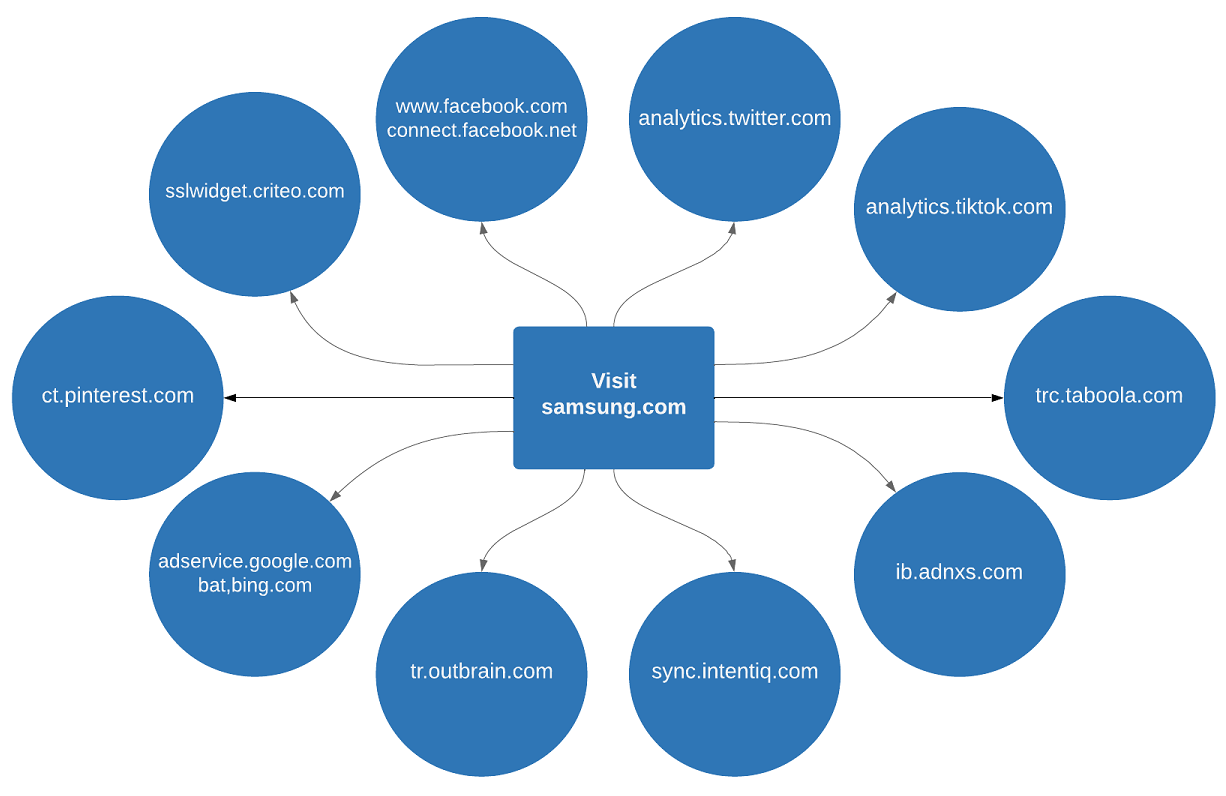}
\caption{External accesses upon visiting samsung.com}
\label{samsung}
\end{figure}

To capture data transmitted to multiple services on the web, one can utilize Mitmproxy by setting up a proxy server that intercepts all communications between a target device and the internet. The process begins with configuring a mobile phone or other internet-connected devices to route their traffic through the proxy server. This setup involves installing Mitmproxy and its associated certificates on the device, allowing the proxy to decipher HTTPS traffic. Once configured, the device communicates with the internet through the proxy server, enabling the capture of requests and responses between the device and various web services.

For instance, a user can visit a website or perform a Google search after connecting a phone to the proxy. Mitmproxy captures all outgoing requests and incoming responses during these interactions, displaying them in real time. This allows researchers to observe the direct communication with the intended web service and subsequent requests sent to third-party services, trackers, and ad networks triggered by the initial user action. The captured data highlights the extent of information being shared with multiple entities, often without the user's explicit awareness, thus providing empirical evidence of the data flows underpinning surveillance capitalism.

Upon visiting \textit{samsung.com} (see Figure~\ref{samsung}), our investigation revealed several external accesses to various services, including but not limited to facebook.com, twitter.com, tiktok.com, and pinterest.com. These services play a distinct role in the digital advertising ecosystem, contributing to a multifaceted approach to online user tracking, profiling, and targeted advertising.

\begin{figure}[ht]
\centering
\includegraphics[width=1\textwidth]{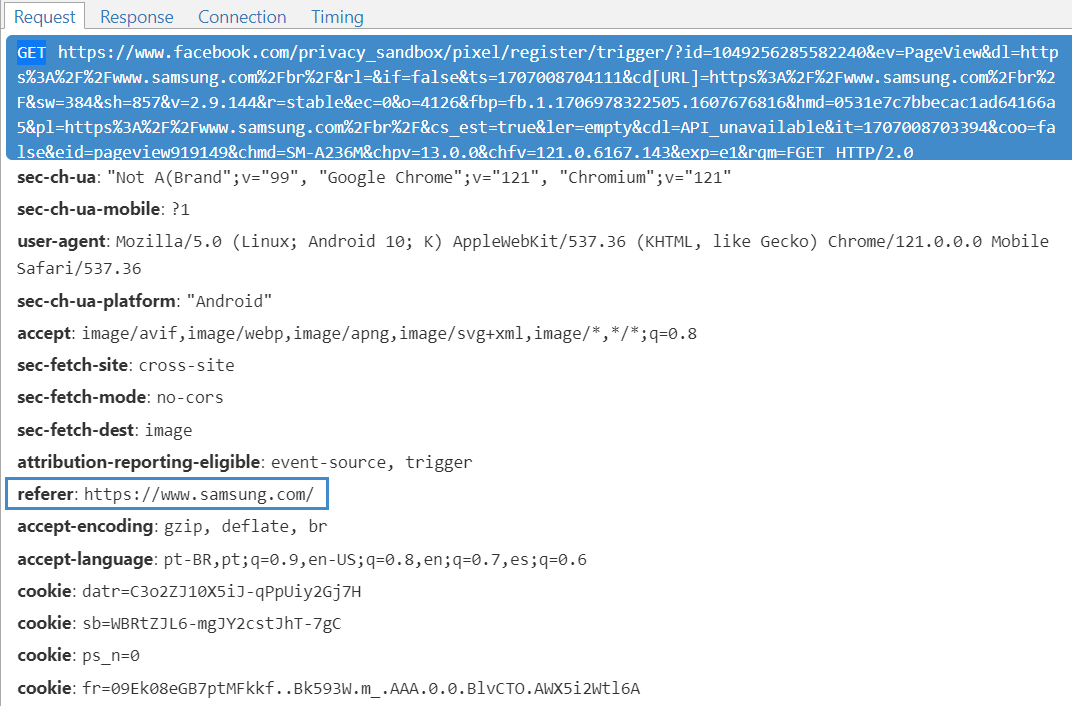}
\caption{Packet payload referring to samsung.com} 
\end{figure}

For instance, Criteo specializes in retargeting, showing ads to users who have visited specific websites, suggesting that visiting \textit{samsung.com} could lead to targeted advertisements from Samsung on other websites. Similarly, Facebook's tracking pixels (via connect.facebook.net and www.facebook.com) allow for the collection of detailed user interactions on Samsung's site, enabling highly personalized advertisements on Facebook platforms. Twitter's analytics services extend this capability into social media engagement, potentially influencing the ads and content users see on Twitter based on their browsing history. TikTok's analytics service is designed to track and analyze user interactions related to TikTok content embedded or shared on that website.

Bing's tracking service (bat.bing.com) and Google's ad services offer insights into user search behavior and preferences, further refining ad targeting capabilities. Taboola and Outbrain specialize in content recommendation, indicating that users might see suggested content related to Samsung products or related interests based on their visit to \textit{samsung.com}.

Adnxs.com (AppNexus) represents a programmatic advertising platform that facilitates real-time bidding for advertising space, suggesting that users' data could be used to auction ad space in real-time to the highest bidder based on the perceived value of the advertising opportunity. Intentiq and similar services (sync.intentiq.com) focus on identity resolution, helping advertisers link activity across devices to a single user, and enhancing cross-device targeting strategies.

Following the capture of network traffic initiated by a visit to \textit{samsung.com}, a subsequent navigation to a news site, like \textit{globo.com}, eventually reveals a significant observation: a series of advertisements originating from \textit{samsung.com} featured across the site. This occurrence directly illustrates the sophisticated mechanisms of online advertising networks and their ability to deliver highly targeted advertisements based on recent user activity. The presence of Samsung advertisements on \textit{globo.com}, shortly after visiting Samsung's official website, shows the efficacy of technologies in tracking user interests and behaviors across the web. 

\subsection{Searching the Web}
This case investigates the surveillance mechanisms search engines activate in response to user queries. This study highlights how search engines collect, store, and possibly share search data by analyzing the network traffic generated from searching specific terms. It explores the potential for profile building based on search history and examines the privacy concerns related to personalized search results and targeted advertising.

\begin{wrapfigure}{r}{0.4\textwidth}
\begin{center}
\includegraphics[width=1\linewidth]
{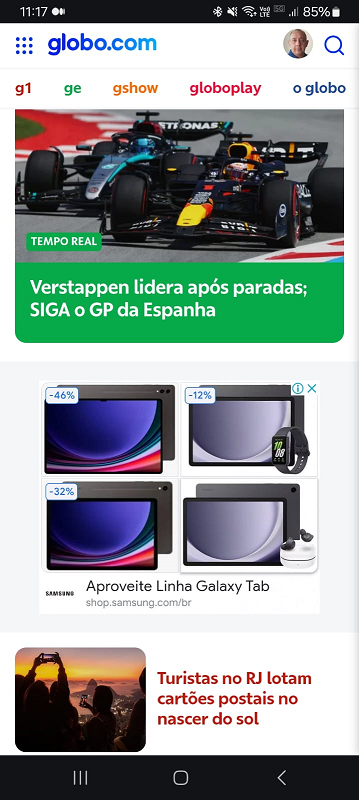}
\end{center}
\caption{Advertising} 
\label{fig:samsung}
\end{wrapfigure}

The privacy concerns arising from these practices are multifaceted. As search engines track and store vast amounts of user data, including search history, location, and personal preferences, they construct detailed user profiles that can be used to serve highly targeted advertisements. This data collection and profiling raise significant privacy issues, as users often remain unaware of the extent of the information being gathered and how it is utilized. There is a lack of transparency, and, worst of all, in our opinion, no user consent is required for these processes.

The captured trace provides a detailed glimpse into the network activities associated with user interactions on various Google platforms. Notably, the trace includes accesses to beacons.gcp.gvt2.com, a domain often utilized for transmitting analytics and performance data, enabling Google to monitor and optimize its services. Additionally, the trace reveals interactions with \url{optimizationguide-pa.googleapis.com}, a critical endpoint for fetching optimization guides and resources that contribute to enhancing the user experience across Google services. These specific accesses highlight the underlying mechanisms involved in data transmission and processing during user engagements with Google's services, showcasing the intricate network architecture and functionalities that facilitate seamless user experiences while facilitating data collection and optimization efforts.

\begin{figure}[ht]
\centering
\includegraphics[width=1\textwidth]{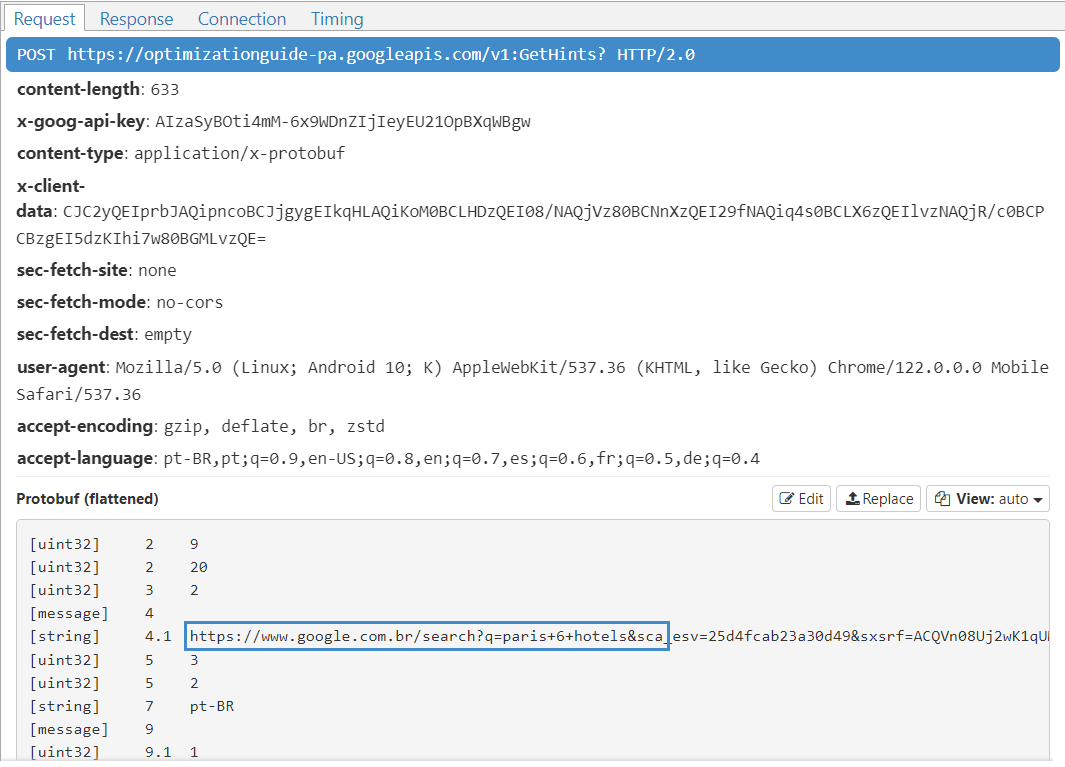}
\caption{Web searching for Paris 6 Hotels} 
\end{figure}

\section{Conclusions and Future Work}
Surveillance capitalism has vividly manifested in web navigation and searching the web, as shown in this study. Users are not just retrieving information from the Web; they're also feeding data into a vast ecosystem that links user data and the digital economy. In our opinion, developing more transparent systems with user consent is imperative.  

A particularly striking observation during our study relates to targeted advertisements appearing on social media platforms shortly after discussing specific topics verbally near a smartphone. Although no direct evidence was collected in this research to confirm the precise mechanisms behind this occurrence, it raises significant concerns about the extent of surveillance capitalism in everyday life. This evidence suggests that voice-activated devices and applications could passively capture audio data, utilizing it for targeted marketing purposes. Future work should aim to investigate and empirically validate these occurrences, exploring whether and how audio data from personal conversations is being harvested and used by web services. Such research would provide critical insights into the invasive nature of surveillance capitalism, highlighting the need for stricter regulations and transparency regarding data collection practices.

Just as Google has taken steps towards transparency in localization services, offering users a clear view of their location history and the option to opt in or out, similar clarity and control must be extended across all data-capturing services. Future work in this field should focus on developing robust mechanisms for obtaining explicit user consent, ensuring that users know the data collected and understand the implications. As we navigate the delicate balance of utility and privacy, empowering users to make informed decisions about their data is paramount to fostering trust and safeguarding the digital ecosystem against practices that may harm individuals and societies.

The challenge remains to find a sustainable model that respects user privacy while leveraging the benefits of data-driven insights. The evolution of Surveillance Capitalism is at a crossroads as businesses and regulators grapple with these complex issues in an increasingly digital world. To achieve a balanced approach between user privacy and the benefits of data-driven insights, the Brazilian government must focus on enhancing the regulatory framework and strengthening the enforcement capabilities of the ANPD (National Data Protection Authority). This involves increasing the ANPD's resources and authority and fostering a culture of compliance through education and awareness campaigns targeting both businesses and the general public. 

The government should also promote the development of privacy-preserving technologies, such as differential privacy and data anonymization techniques, to enable businesses to extract valuable insights from data without compromising individual privacy. Additionally, creating more explicit guidelines on the application of legitimate interests as a basis for data processing would help harmonize the goals of innovation and privacy protection. Finally, encouraging open dialogue between stakeholders, regulators, businesses, civil society, and academia- can facilitate the creation of policies that adapt to technological advances while safeguarding citizens' rights, ensuring that data-driven growth does not come at the expense of fundamental privacy protections.


\bibliographystyle{sbc}
\bibliography{surveillance_capitalism}

\end{document}